\documentclass[10pt,twocolumn,letterpaper]{article}

\usepackage[pagenumbers]{cvpr} 

\usepackage{cuted} 

\usepackage{currfile} 

\usepackage{caption} 
\definecolor{cvprblue}{rgb}{0.21,0.49,0.74}
\usepackage[pagebackref,breaklinks,colorlinks,allcolors=cvprblue]{hyperref}
\def\confName{CVPR}
\def\confYear{2027}

\title{Beyond Pairwise Similarity: Efficient Token Compression via Spatiotemporal Basis Construction in Video Large Language Models}

\author{
Jiale Xin$^{1}$,
Ke Li$^{1}$\thanks{Corresponding authors},
Jiayi Wang$^{1}$,
Di Wang$^{1}$\footnotemark[1],
Weiping Ni$^{1}$,
Tao Lei$^{2}$,
Yiming Zhang$^{3}$,
Quan Wang$^{1}$\\
$^1$Xidian University \enspace $^2$Fuzhou University \enspace $^3$University of California San Diego \\
}

\usepackage{booktabs}
\usepackage{amssymb}
\usepackage{multirow}
\usepackage{makecell}
\usepackage{graphicx}
\usepackage{colortbl}
\usepackage{stfloats}

\definecolor{oursgray}{gray}{0.92}
\newcommand{\ourscell}[1]{\cellcolor{oursgray}#1}
\begin{document}

\maketitle
\begin{abstract}
Although video large language models (VLLMs) demonstrate
remarkable video understanding capabilities,  excessive visual tokens remain a bottleneck for long-video inference. Existing compression methods typically prune or merge visual tokens based on token similarity. However, is similarity alone sufficient to identify redundancy? 
In this work, we find that some low-similarity tokens can still be jointly reconstructed from other tokens, indicating that low similarity does not necessarily imply low redundancy. Inspired by sparse representation, in which a signal can be reconstructed from a few basis vectors, we propose MTRC, a training-free compression framework for VLLMs. MTRC constructs a compact spatiotemporal basis set that represents as much video information as possible. 
Specifically, a Temporal Selection Module (TSM) is designed to reconstruct the current frame using tokens retained from previous frames and add difficult-to-reconstruct tokens to a candidate pool.
A Spatial Filtering Module (SFM) is further introduced to jointly reconstruct candidate tokens using a few same-frame tokens and remove redundant content that can be jointly represented.
Through the synergy of the two modules, MTRC achieves more efficient video compression.
Extensive experiments across four benchmarks and two VLLM backbones validate both the effectiveness and generalizability of MTRC. 
Notably, on LLaVA-OneVision, MTRC retains only 15\% of visual tokens while achieving 100.2\% of the uncompressed model's average accuracy and delivering a $5.7\times$ speedup in LLM prefilling.
\end{abstract}
\begin{figure}[t]
    \centering
    \includegraphics[width=\linewidth]{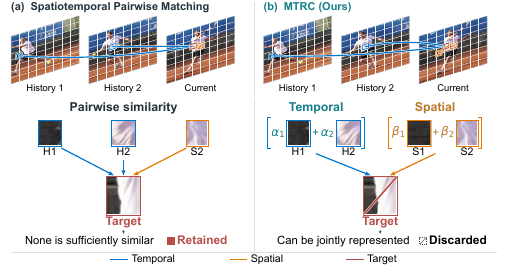}
\caption{Comparison of token redundancy assessment methods.
(a) Pairwise similarity retains the target token because no individual reference is sufficiently similar.
(b) MTRC uses multi-token reconstruction to represent the target feature with historical or spatial tokens and removes the target token as redundant.}
    \label{fig:motivation}
\end{figure}

\section{Introduction}
\label{sec:intro}

In recent years, video large language models (VLLMs) have demonstrated remarkable video understanding capabilities~\cite{li2025llavaonevisioneasyvisualtasktransfer,zhang2025llavavideo}.
These models typically encode sampled video frames into visual tokens, which are then fed into a language model for understanding and reasoning.
However, the computational complexity of attention grows quadratically with sequence length.
Processing large numbers of visual tokens therefore significantly increases inference overhead~\cite{dao2024flashattention,shen2026fastvid,fan2026flashvid}.
To address this challenge, token compression has emerged as a promising approach to efficient inference, substantially reducing computation by pruning or merging redundant tokens~\cite{chen2024image,yang2025visionzip,shen2026fastvid,bolya2022token,huang2025prunevid,fan2026flashvid}.

Most existing video token compression methods assess redundancy based on pairwise token similarity: when two tokens are sufficiently similar, one is considered redundant and removable.
This strategy has been effective for token compression along either the spatial or temporal dimension~\cite{shang2025llava,yang2025visionzip,wang2025folder,fu2025framefusion}.
However, videos exhibit complex spatial structures and dynamic temporal changes.
Removing redundancy along only one dimension cannot fully exploit these spatiotemporal relationships.
Recent studies have therefore explored joint spatiotemporal compression to identify and remove redundant tokens along both dimensions~\cite{huang2025prunevid,shao2026holitom,fan2026flashvid}.
Although these methods account for video redundancy more comprehensively, their redundancy assessment still primarily relies on pairwise token similarity.
As shown in Fig.~\ref{fig:motivation}(a), the token at the boundary between the clothing and the background has low similarity to tokens inside either region.
It is therefore misidentified as irreplaceable and retained.

However, \textit{does the absence of a sufficiently similar individual token mean that a token is irreplaceable?}
Figure~\ref{fig:motivation}(b) illustrates another possibility for the same boundary token retained in Fig.~\ref{fig:motivation}(a).
This token can still be approximated by a linear combination of clothing and background tokens from the same and other frames.
This indicates that redundancy depends fundamentally on whether a token's information can be jointly represented by a group of tokens, rather than on how close two tokens are.
\textbf{We therefore shift the focus of video token compression from pairwise token proximity to the overall coverage of the constructed set.}
This perspective aligns with the sparse representation principle of approximating high-dimensional signals through linear combinations of a small number of basis elements~\cite{mairal2010online,elhamifar2012see}.

Existing methods for compressing vision-language models (VLMs) use a small set of tokens to approximate the original visual tokens, thereby accelerating VLMs~\cite{ma2026apet}.
However, extending this idea to video requires basis selection to account for relationships both within and across frames.
Repeated spatial structures within a frame lead to similar information in neighboring regions.
Temporal continuity across frames causes the same content to recur in consecutive frames~\cite{huang2025prunevid,tao2025dycoke}.
Building on this observation, we revisit video token compression, aiming to represent as much video information as possible with a compact spatiotemporal basis set.

Motivated by these observations, we propose MTRC, a training-free token compression framework for VLLMs.
It uses reconstruction error to measure the joint representational capacity of multiple tokens.
Specifically, we propose the Temporal Selection Module (TSM), which reconstructs the current frame using tokens retained from previous frames.
It selects difficult-to-reconstruct tokens for inclusion in a candidate pool, thereby reducing the repeated retention of existing information across frames.
However, tokens in the candidate pool may still be redundant within the same frame.
To this end, the Spatial Filtering Module (SFM) reconstructs candidate tokens using a small number of same-frame tokens, effectively removing this redundancy.
Finally, through coordinated selection by these two modules, MTRC achieves efficient compression that adapts to video content.

We systematically evaluate MTRC on four video understanding benchmarks~\cite{fu2025video,mangalam2023egoschema,wu2024longvideobench,li2024mvbench} using two representative vision-language models, LLaVA-OneVision~\cite{li2025llavaonevisioneasyvisualtasktransfer} and LLaVA-Video~\cite{zhang2025llavavideo}.
Notably, on LLaVA-OneVision, MTRC retains only 15\% of visual tokens while achieving 100.2\% of the uncompressed model's average performance across the four benchmarks.
It also achieves a $5.7\times$ speedup in LLM prefilling and a $1.9\times$ speedup in time to first token (TTFT).
These results demonstrate that MTRC reduces inference costs while maintaining video understanding performance.

To summarize, our main contributions are as follows:
\begin{itemize}
    \item We reveal that pairwise similarity alone is insufficient to fully identify token redundancy. Building on this observation, we revisit video token compression from the perspective of spatiotemporal basis selection, shifting the focus to the information coverage of the basis set and offering a new perspective on token compression.

    \item We propose MTRC, a training-free token compression method for VLLMs that combines the Temporal Selection Module (TSM) with the Spatial Filtering Module (SFM) to construct a compact spatiotemporal basis set, thereby efficiently reducing video redundancy.

    \item Extensive experiments demonstrate that MTRC reduces inference costs with negligible performance degradation and generalizes well across different models and evaluation settings.
\end{itemize}
\section{Related Work}
\label{sec:related_work}
\subsection{Video Large Language Models}
\label{sec:video_llms}

In recent years, multimodal large language models have expanded their scope from static images to video inputs, supporting video captioning, reasoning, and question answering~\cite{maaz2024video,zhang2023video,lin2024video,wang2024qwen2,shen2024longvu,cheng2024videollama,zhang2025videollama}.
Models such as LLaVA-OneVision~\cite{li2025llavaonevisioneasyvisualtasktransfer}, LLaVA-Video~\cite{zhang2025llavavideo}, and Qwen3-VL~\cite{bai2025qwen3} typically combine a visual encoder, a cross-modal connector, and a large language model.
This architecture aligns video tokens with language representations to support cross-modal understanding.
These models have demonstrated strong performance across multiple video understanding benchmarks.

Despite their impressive video understanding capabilities, increasing the number of input frames lengthens the visual sequence, imposing higher computational and memory costs on the language model~\cite{shen2026fastvid,fan2026flashvid,song2024moviechat}.
This issue becomes even more severe for long-video understanding~\cite{shu2025video}.
Therefore, compressing visual inputs while preserving existing model capabilities has become an important direction for improving inference efficiency, motivating growing research on visual token compression~\cite{cha2024honeybee,arif2025hired,xing2025conical,zhang2025llava,tao2025dycoke,shen2024longvu}.

\subsection{Visual Token Compression}
\label{sec:visual_token_compression}

Many visual token compression methods use attention or similarity to guide pruning or merging, shorten visual sequences and reduce inference costs~\cite{shang2025llava,alvar2025divprune,zhang2025beyond,liu2024multi}.
For example, FastV~\cite{chen2024image} prunes visual tokens using attention from the early layers of the language model, reducing computation in subsequent layers.
FOLDER~\cite{wang2025folder} matches and merges similar tokens within individual frames to reduce spatial redundancy.
In contrast, ApET~\cite{ma2026apet} selects representative tokens
based on approximation error and has been applied to both
image and video understanding tasks. However, its compression
strategy does not explicitly exploit temporal redundancy
across frames.
These studies provide different scoring criteria for visual token compression.

Recent studies have explored video token compression
from a spatiotemporal perspective~\cite{shao2025survey}.
PruneVid~\cite{huang2025prunevid} merges temporally static and spatially similar tokens and performs question-guided pruning within the language model.
FastVID~\cite{shen2026fastvid} combines dynamic temporal segmentation, density-driven merging, and attention-driven selection.
HoliTom~\cite{shao2026holitom} combines temporal segmentation, spatial merging, and merging within the language model.
EarlyTom~\cite{wang2026earlytom} moves temporal merging into the vision encoder and adopts different spatial selection strategies for dynamic and static content.
FlashVID~\cite{fan2026flashvid} combines attention- and diversity-based selection with token merging over cross-frame redundancy trees.
However, these methods primarily score tokens based on attention or pairwise similarity, without directly assessing whether a token can be jointly represented by multiple tokens.
Consequently, some redundant tokens that could be jointly represented by multiple tokens may still be retained, consuming the limited token budget.

\begin{figure*}[t]
    \centering
    \includegraphics[width=\textwidth]{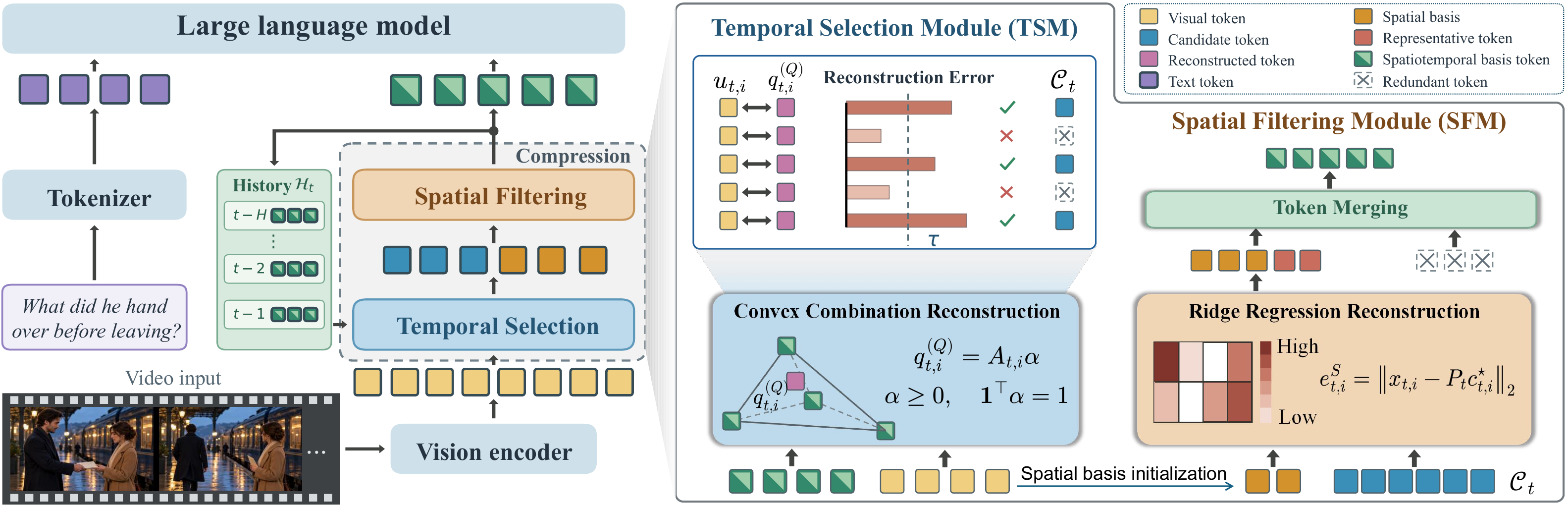}
    \caption{
Our MTRC compresses visual tokens by two complementary modules: 
Temporal Selection Module (TSM) reconstructs current-frame tokens from the historical basis, adding high-error tokens to a candidate pool;
Spatial Filtering Module (SFM) reconstructs candidates from a same-frame spatial basis and selects representatives by reconstruction error.
After token merging, the representatives and spatial basis form the current frame's spatiotemporal basis set, which updates historical memory to provide references for subsequent frames.
}
    \label{fig:framework}
\end{figure*}

\section{Method}

\subsection{Overview}

As shown in Fig.~\ref{fig:framework}, MTRC integrates two synergistic modules for spatiotemporal basis selection.
The Temporal Selection Module (TSM) reduces the repeated retention of existing information across frames, allocating more of the budget to changes in video content.
The Spatial Filtering Module (SFM) selects representatives that
supplement information not adequately represented by a shared
same-frame spatial basis.

\subsection{Temporal Selection Module}

Consecutive frames contain both persistent content and new information introduced by object motion and scene changes.
To determine which information in the current frame still needs to be retained, we propose the Temporal Selection Module (TSM).
It reconstructs current-frame tokens using combinations of historical basis tokens and places difficult-to-reconstruct tokens into a candidate pool for further selection by the subsequent Spatial Filtering Module (SFM).

\noindent\textbf{Rolling memory.}
The rolling memory stores the compressed outputs of previous frames, providing a historical basis set for the temporal reconstruction of the current frame.
Let $\mathcal{H}_t$ denote the rolling memory before processing frame $t$.
For subsequent frames, the rolling memory consists of the compressed outputs of at most the $H$ most recent sampled frames:
\begin{equation}
\mathcal{H}_t
=
\operatorname{Concat}\!\left(
Y_{\max(1,t-H)},\ldots,Y_{t-1}
\right),
\qquad t>1.
\label{eq:strc_memory}
\end{equation}
Here, $H$ is the memory window length, and $Y_t$ is the final compressed output of frame $t$.
After each frame is compressed, its output is added to the rolling memory.
When the window length is exceeded, the output of the oldest frame is removed.
Therefore, TSM uses cross-frame representations that have actually been retained.
This allows current-frame token selection to account for existing compression results and significantly reduces temporal redundancy at low retention ratios.

\noindent\textbf{Temporal reconstruction scoring.}
Let $X_t=[x_{t,1},\ldots,x_{t,N}]\in\mathbb{R}^{d\times N}$
denote the patch-token features of frame $t$, where $N$
is the number of patch tokens and $d$ is the feature dimension.
For a current token $x_{t,i}$ and a historical basis token $h_j$,
their normalized features are defined as
$u_{t,i}=x_{t,i}/\|x_{t,i}\|_2$
and
$v_j=h_j/\|h_j\|_2$, respectively.
Based on the cosine similarity $u_{t,i}^{\top}v_j$, we retrieve the $B$ most similar historical basis tokens.
The retrieved basis tokens are ordered by similarity to form a token-specific temporal dictionary $A_{t,i}\in\mathbb{R}^{d\times B}$, determining which historical basis tokens are used.
The columns of $A_{t,i}$ are the retrieved normalized historical features $v_j$, arranged in the same descending order of similarity.
We evaluate the temporal reconstructability of the current token through the following simplex-constrained optimization problem:
\begin{equation}
\begin{aligned}
&\min_{\alpha\in\Delta_B}
\frac{1}{2}\left\|u_{t,i}-A_{t,i}\alpha\right\|_2^2,\\
&\Delta_B=
\left\{
\alpha\in\mathbb{R}^{B}:
\alpha\geq 0,\;
\mathbf{1}^{\top}\alpha=1
\right\}.
\end{aligned}
\label{eq:strc_simplex}
\end{equation}

The historical basis tokens form a historical dictionary for the current token.
Unconstrained linear reconstruction allows extrapolation and may
involve coefficients with large magnitudes.
We constrain the coefficients to the simplex, restricting reconstruction
to convex combinations of the retrieved historical tokens.
The non-negative, sum-to-one weights also limit coefficient-induced
amplification of errors in the history.
We approximately solve Eq.~\eqref{eq:strc_simplex} using $Q$ Frank--Wolfe iterations~\cite{frank1956algorithm,jaggi2013revisiting}. For clarity, we omit the token indices $(t,i)$ in the following iterative updates.
Let $a_j$ denote the $j$-th column of the dictionary.
We initialize the reconstructed feature as $q^{(0)}=a_1$,
where $a_1$ is the normalized historical feature most similar to the current token.
At each iteration, we compute
\begin{equation}
j_r=
\arg\min_j a_j^\top(q^{(r)}-u),
\qquad
d^{(r)}=a_{j_r}-q^{(r)},
\label{eq:strc_fw_direction}
\end{equation}
\begin{equation}
\begin{aligned}
&\gamma_r
=
\begin{cases}
\displaystyle
\operatorname{clip}_{[0,1]}
\left(
\frac{(u-q^{(r)})^\top d^{(r)}}
{\|d^{(r)}\|_2^2}
\right),
& \|d^{(r)}\|_2^2>\varepsilon^2\\
0
& \text{otherwise},
\end{cases},\\
&q^{(r+1)}
=
q^{(r)}+\gamma_r d^{(r)},
\end{aligned}
\label{eq:strc_fw_update}
\end{equation}
where $q^{(r)}$ is the reconstructed feature at iteration $r$,
$u-q^{(r)}$ is the current residual, and
$\varepsilon>0$ is a numerical stability constant.
Equation~\eqref{eq:strc_fw_direction} selects the dictionary column
that minimizes the first-order approximation of the objective function.
The reconstructed feature is then updated with a step size $\gamma_r\in[0,1]$, ensuring that it remains a convex combination of historical basis tokens.

After $Q$ iterations, we use the Euclidean distance between the current token and the final reconstructed feature as the temporal score:
\begin{equation}
e^{\mathrm{T}}_{t,i}
=
\|u_{t,i}-q^{(Q)}_{t,i}\|_2.
\label{eq:strc_temporal_score}
\end{equation}

This score measures the approximate reconstructability of the current token under the historical dictionary.
A higher score indicates that, even after multiple iterations, the historical basis set still cannot adequately reconstruct the current feature.
We use this score to assess temporal redundancy and place tokens with larger reconstruction errors into the candidate pool, thereby reducing temporal redundancy.

\noindent\textbf{Threshold-gated candidate pool.}
TSM and SFM are connected through the candidate pool.
The former reduces temporal redundancy, whereas the latter further
reduces spatial redundancy on this basis, jointly accomplishing
spatiotemporal basis selection.

Let $\mathcal{I}=\{1,\ldots,N\}$ denote the index set of all patch
tokens in the current frame.
For the first frame, we bypass TSM and use all patch tokens
as candidates, setting $\mathcal{C}_1=\mathcal{I}$.
For subsequent frames ($t>1$), the initial candidate set is defined as
\begin{equation}
\mathcal{C}_t^{(0)}
=
\mathcal{S}_t
\cup
\left\{
i\in\mathcal{I}:
e^{\mathrm{T}}_{t,i}>\tau
\right\},
\label{eq:strc_candidate_pool}
\end{equation}
where $\tau$ is the temporal filtering threshold, and
$\mathcal{S}_t$ is the spatial basis index set.
The spatial basis is exempt from temporal gating and retained
within the final budget.
This safeguards current-frame information coverage while reducing
the dependence of subsequent selection on previously compressed
representations.
The spatial basis selection procedure is described in Section~3.3.

To leave room for subsequent spatial selection, we set the minimum candidate count to a prescribed multiple of the final output budget.
If thresholding produces too few candidates, we add unselected tokens in descending order of temporal score until this minimum is reached.
Otherwise, all tokens satisfying the threshold are retained. We denote the resulting candidate index set after this completion step by $\mathcal C_t$.
In this way, the candidate pool size can vary with video content, and the spatial stage performs further selection to meet the fixed output budget.

\subsection{Spatial Filtering Module}

After temporal selection, some candidate tokens may still be well
represented by a small set of same-frame tokens.
We propose the Spatial Filtering Module (SFM) to select representatives
that complement this shared spatial basis.

\noindent\textbf{Spatial basis initialization.}
To provide reference anchors for subsequent spatial reconstruction, we first select $S$ tokens from the original features $X_t$ of each frame,
forming the spatial basis index set $\mathcal{S}_t$.
\begin{table*}[t]
\centering
\caption{
Performance comparison of token compression methods on LLaVA-OV-7B.
}
\label{tab:strc_main}

\begingroup

\resizebox{\textwidth}{!}{%
\begin{tabular}{@{}l*{11}{c}@{}}
\toprule

\multirow{2}{*}{Method}
& \multirow{2}{*}{\makecell{Retention\\$R$}}
& \multicolumn{4}{c}{VideoMME}
& \multicolumn{2}{c}{EgoSchema}
& \multirow{2}{*}{\makecell{LongVideo\\Bench}}
& \multirow{2}{*}{MVBench}
& \multirow{2}{*}{\makecell{Avg.\\Score}}
& \multirow{2}{*}{\makecell{Rel.\ Acc.\\(\%)}} \\

\cmidrule(lr){3-6}
\cmidrule(lr){7-8}

& & Short & Medium & Long & Overall
& Subset & Total
& & & & \\

\midrule

LLaVA-OV-7B
& 100\%
& 69.8 & 56.8 & 48.7 & 58.4
& 62.2 & 60.2 & 56.6 & 58.3 & 58.4 & 100.0 \\

\midrule


VisionZip~\cite{yang2025visionzip} \textcolor{gray}{(CVPR'25)}
& \multirow{6}{*}{25\%}
& 68.8 & \textbf{57.4} & 48.1 & \underline{58.1}
& \underline{62.2} & \underline{60.3}
& 56.3 & \textbf{57.6} & \underline{58.1} & \underline{99.5} \\

PruneVid~\cite{huang2025prunevid} \textcolor{gray}{(ACL Findings'25)}
&
& 67.4 & 54.9 & 47.0 & 56.4
& 61.4 & 58.0
& 55.7 & 56.8 & 56.7 & 97.2 \\

FastVID~\cite{shen2026fastvid} \textcolor{gray}{(NeurIPS'25)}
&
& 69.9 & 56.2 & 47.4 & 57.8
& 61.2 & 58.2
& 55.9 & 56.5 & 57.1 & 97.8 \\

EarlyTom~\cite{wang2026earlytom} \textcolor{gray}{(CVPR'26)}
&
& 69.4 & \underline{56.4} & \underline{48.3} & 58.0
& 61.8 & \textbf{60.5}
& 56.3 & \underline{57.4} & \underline{58.1} & 99.4 \\

FlashVID~\cite{fan2026flashvid} \textcolor{gray}{(ICLR'26)}
&
& \underline{70.1} & 55.9 & 48.1 & 58.0
& \underline{62.2} & 60.1
& \underline{56.7} & \textbf{57.6} & \underline{58.1} & \underline{99.5} \\

\textbf{MTRC (Ours)}
&
& \ourscell{\textbf{71.5}}
& \ourscell{56.2}
& \ourscell{\textbf{48.5}}
& \ourscell{\textbf{58.7}}
& \ourscell{\textbf{62.4}}
& \ourscell{\underline{60.3}}
& \ourscell{\textbf{57.4}}
& \ourscell{\textbf{57.6}}
& \ourscell{\textbf{58.5}}
& \ourscell{\textbf{100.2}} \\

\midrule


VisionZip~\cite{yang2025visionzip} \textcolor{gray}{(CVPR'25)}
& \multirow{6}{*}{20\%}
& 68.4 & \textbf{57.0} & 48.1 & 57.8
& 62.0 & 59.8
& 55.2 & \underline{57.6} & 57.6 & 98.7 \\

PruneVid~\cite{huang2025prunevid} \textcolor{gray}{(ACL Findings'25)}
&
& 67.2 & 53.9 & \underline{48.2} & 56.4
& \underline{63.2} & 59.7
& 54.7 & 56.2 & 56.8 & 97.2 \\

FastVID~\cite{shen2026fastvid} \textcolor{gray}{(NeurIPS'25)}
&
& 69.9 & 56.3 & 47.2 & 57.8
& 61.0 & 57.9
& 55.6 & 56.3 & 56.9 & 97.5 \\

EarlyTom~\cite{wang2026earlytom} \textcolor{gray}{(CVPR'26)}
&
& 69.2 & \underline{56.5} & 48.0 & \underline{57.9}
& 62.0 & \underline{60.4}
& 55.6 & 57.5 & \underline{57.9} & 99.1 \\

FlashVID~\cite{fan2026flashvid} \textcolor{gray}{(ICLR'26)}
&
& \underline{70.1} & 54.5 & 47.9 & 57.5
& 62.8 & 60.1
& \underline{56.2} & \textbf{57.8} & \underline{57.9} & \underline{99.2} \\

\textbf{MTRC (Ours)}
&
& \ourscell{\textbf{70.3}}
& \ourscell{55.0}
& \ourscell{\textbf{49.1}}
& \ourscell{\textbf{58.1}}
& \ourscell{\textbf{63.4}}
& \ourscell{\textbf{60.6}}
& \ourscell{\textbf{56.9}}
& \ourscell{\underline{57.6}}
& \ourscell{\textbf{58.3}}
& \ourscell{\textbf{99.9}} \\

\midrule


VisionZip~\cite{yang2025visionzip} \textcolor{gray}{(CVPR'25)}
& \multirow{7}{*}{15\%}
& 63.2 & 54.6 & 48.5 & 55.4
& \underline{62.6} & 59.8
& 54.3 & 53.4 & 55.7 & 95.5 \\

PruneVid~\cite{huang2025prunevid} \textcolor{gray}{(ACL Findings'25)}
&
& 67.5 & 52.4 & 46.4 & 55.4
& 61.6 & 57.5
& 54.8 & 55.4 & 55.8 & 95.5 \\

FastVID~\cite{shen2026fastvid} \textcolor{gray}{(NeurIPS'25)}
&
& 68.3 & \underline{55.0} & 47.4 & 56.9
& 59.6 & 57.4
& 56.2 & 56.0 & 56.6 & 97.0 \\

EarlyTom~\cite{wang2026earlytom} \textcolor{gray}{(CVPR'26)}
&
& 68.6 & 54.8 & 47.3 & 56.9
& 60.0 & \underline{60.2}
& 54.4 & \underline{57.5} & 57.3 & 98.1 \\
FlashVID~\cite{fan2026flashvid} \textcolor{gray}{(ICLR'26)}
&
& \underline{69.2} & \underline{55.0}
& \underline{48.9} & \underline{57.7}
& \underline{62.6} & \textbf{60.3}
& \underline{56.5} & 57.3
& \underline{58.0} & \underline{99.3} \\

ApET~\cite{ma2026apet} \textcolor{gray}{(CVPR'26)}
&
& 68.9 & 53.7 & \underline{48.9} & 57.2
& 62.2 & 59.5
& 56.2 & 56.3 & 57.3 & 98.1 \\

\textbf{MTRC (Ours)}
&
& \ourscell{\textbf{69.9}}
& \ourscell{\textbf{55.2}}
& \ourscell{\textbf{50.3}}
& \ourscell{\textbf{58.5}}
& \ourscell{\textbf{62.8}}
& \ourscell{\textbf{60.3}}
& \ourscell{\textbf{57.5}}
& \ourscell{\textbf{57.6}}
& \ourscell{\textbf{58.5}}
& \ourscell{\textbf{100.2}} \\

\midrule


VisionZip~\cite{yang2025visionzip} \textcolor{gray}{(CVPR'25)}
& \multirow{6}{*}{10\%}
& 60.2 & 52.8 & 46.7 & 53.2
& 61.6 & 58.0
& 49.3 & 53.5 & 53.5 & 91.6 \\

PruneVid~\cite{huang2025prunevid} \textcolor{gray}{(ACL Findings'25)}
&
& 65.9 & 52.8 & 45.5 & 54.7
& 60.2 & 57.2
& 54.0 & 53.7 & 54.9 & 94.0 \\

FastVID~\cite{shen2026fastvid} \textcolor{gray}{(NeurIPS'25)}
&
& \underline{67.6} & \textbf{55.4} & \underline{48.4} & \textbf{57.1}
& 58.8 & 56.5
& \underline{55.3} & 55.9 & 56.2 & 96.3 \\

EarlyTom~\cite{wang2026earlytom} \textcolor{gray}{(CVPR'26)}
&
& 66.2 & 53.8 & 47.5 & 55.8
& 59.0 & \underline{59.2}
& 52.4 & \underline{56.5} & 56.0 & 95.9 \\

FlashVID~\cite{fan2026flashvid} \textcolor{gray}{(ICLR'26)}
&
& 67.5 & 53.9 & \textbf{48.7} & \underline{56.7}
& \underline{61.8} & 59.1
& 54.2 & 56.3 & \underline{56.6} & \underline{96.9} \\

\textbf{MTRC (Ours)}
&
& \ourscell{\textbf{67.7}}
& \ourscell{\underline{54.8}}
& \ourscell{\textbf{48.7}}
& \ourscell{\textbf{57.1}}
& \ourscell{\textbf{62.4}}
& \ourscell{\textbf{59.5}}
& \ourscell{\textbf{55.6}}
& \ourscell{\textbf{56.9}}
& \ourscell{\textbf{57.3}}
& \ourscell{\textbf{98.1}} \\

\bottomrule
\end{tabular}%
}

\endgroup

\smallskip

\end{table*}
The first frame has no historical reference, and its compressed
output is used to initialize the rolling memory.
Farthest point sampling (FPS)~\cite{qi2017pointnet++} primarily
expands feature coverage based on distance.
In contrast, we additionally consider the local density of basis
points in the first frame, aiming to balance the feature
representativeness and separation of the initial spatial basis.
Specifically, DPC-KNN~\cite{zeng2022not} is used for the first frame,
and FPS is used for subsequent frames, and farthest point sampling (FPS)~\cite{qi2017pointnet++} is used for subsequent frames:
\begin{equation}
\mathcal{S}_t=
\begin{cases}
\operatorname{DPC\text{-}KNN}(X_t,S),
& t=1,\\
\operatorname{FPS}(X_t,S),
& t>1.
\end{cases}
\label{eq:strc_basis_selection}
\end{equation}
Here, $S$ denotes the spatial basis size of the current frame,
with the first-frame and subsequent-frame settings specified
in Appendix~C.

\noindent\textbf{Spatial reconstruction scoring.}
Unlike temporal reconstruction, which retrieves a historical dictionary for each token,
SFM uses a small number of current-frame tokens to form a spatial basis shared across the entire frame.
Because this shared basis is not selected separately for each candidate, we adopt ridge regression~\cite{hoerl1970ridge} to allow more flexible linear combinations.
Regularization is also used to suppress excessively large coefficients.
We preferentially retain candidates with larger reconstruction errors to supplement information not adequately represented by the shared spatial basis.

Let $\mathcal S_t=\{b_1,\ldots,b_S\}$ denote the spatial
basis index set, and let
$P_t=[x_{t,b_1},\ldots,x_{t,b_S}]\in\mathbb{R}^{d\times S}$
denote the corresponding shared spatial basis matrix.
\begin{equation}
\begin{aligned}
c^\star_{t,i}
&=
\arg\min_{c\in\mathbb{R}^{S}}
\left\{
\|x_{t,i}-P_tc\|_2^2
+
\lambda\|c\|_2^2
\right\}\\
&=
(P_t^\top P_t+\lambda I_S)^{-1}
P_t^\top x_{t,i},
\end{aligned}
\label{eq:strc_spatial_ridge}
\end{equation}
Here, $\lambda>0$ is the regularization coefficient, and
$I_S$ is the $S\times S$ identity matrix.
The spatial score of a candidate token is defined as
\begin{equation}
e^{\mathrm{S}}_{t,i}
=
\left\|
x_{t,i}-P_tc^\star_{t,i}
\right\|_2,
\label{eq:strc_spatial_score}
\end{equation}
where $x_{t,i}$ is the current candidate feature, and
$P_t c^\star_{t,i}$ is its reconstructed representation.
We use this reconstruction error to measure spatial redundancy.

Let $K_t$ denote the final output token budget for frame $t$.
The $S$ spatial basis tokens are included in this budget.
We select the $K_t-S$ representatives with the highest spatial
reconstruction scores $e^{\mathrm{S}}_{t,i}$ from
$\mathcal{C}_t\setminus\mathcal{S}_t$, excluding the retained
spatial basis tokens from the selection.

\noindent\textbf{Token merging.}
To reduce information loss caused by directly discarding tokens, we first use the selected representative tokens as group centers.
Each remaining original token is assigned to the representative with the highest similarity, forming groups of tokens.
After grouping, we average the token features within each group to obtain the merged representative tokens.
Finally, the merged representative tokens and the unchanged spatial basis jointly form the spatiotemporal basis set of the current frame,
which is fed into the large language model as visual input.

\begin{table*}[t]
\centering
\caption{
Performance comparison of visual token compression methods on LLaVA-Video.
}
\label{tab:llava_video_main_provisional}

\begingroup
\providecommand{\ourscell}[1]{\cellcolor[gray]{0.92}#1}

\resizebox{\textwidth}{!}{%
\begin{tabular}{@{}l*{11}{c}@{}}
\toprule

\multirow{2}{*}{Method}
& \multirow{2}{*}{\makecell{Retention\\$R$}}
& \multicolumn{4}{c}{VideoMME}
& \multicolumn{2}{c}{EgoSchema}
& \multirow{2}{*}{\makecell{LongVideo\\Bench}}
& \multirow{2}{*}{MVBench}
& \multirow{2}{*}{\makecell{Avg.\\Score}}
& \multirow{2}{*}{\makecell{Rel.\ Acc.\\(\%)}} \\

\cmidrule(lr){3-6}
\cmidrule(lr){7-8}

& & Short & Medium & Long & Overall
& Subset & Total
& & & & \\

\midrule

LLaVA-Video
& 100\%
& 77.0 & 61.8 & 53.6 & 64.1
& 59.4 & 57.3
& 58.6 & 61.0 & 60.3 & 100.0 \\

\midrule


VisionZip~\cite{yang2025visionzip} \textcolor{gray}{(CVPR'25)}
& \multirow{4}{*}{25\%}
& \underline{74.2} & \textbf{60.3}
& \underline{52.1} & \underline{62.2}
& \textbf{58.2} & \textbf{56.8}
& 58.3 & 60.0
& \underline{59.3} & \underline{98.5} \\

FastVID~\cite{shen2026fastvid} \textcolor{gray}{(NeurIPS'25)}
&
& 74.0 & 60.1 & 51.8 & 62.0
& 57.4 & 55.4
& 58.2 & \textbf{60.2} & 58.9 & 97.8 \\

FlashVID~\cite{fan2026flashvid} \textcolor{gray}{(ICLR'26)}
&
& 74.1 & \underline{60.2} & 52.0 & 62.1
& \textbf{58.2} & 56.1
& \underline{58.7} & \underline{60.1}
& \underline{59.3} & 98.3 \\

\textbf{MTRC (Ours)}
&
& \ourscell{\textbf{74.7}}
& \ourscell{59.8}
& \ourscell{\textbf{53.4}}
& \ourscell{\textbf{62.6}}
& \ourscell{\underline{58.0}}
& \ourscell{\underline{56.2}}
& \ourscell{\textbf{59.2}}
& \ourscell{\textbf{60.2}}
& \ourscell{\textbf{59.6}}
& \ourscell{\textbf{98.8}} \\

\midrule


VisionZip~\cite{yang2025visionzip} \textcolor{gray}{(CVPR'25)}
& \multirow{4}{*}{20\%}
& 72.2 & 59.2 & \underline{52.1} & 61.2
& \textbf{59.0} & \underline{56.7}
& 58.1 & 59.9 & 59.0 & 97.9 \\

FastVID~\cite{shen2026fastvid} \textcolor{gray}{(NeurIPS'25)}
&
& \underline{74.1} & \textbf{60.8}
& 51.6 & \underline{62.2}
& 57.8 & 55.6
& 57.7 & \underline{60.2} & 58.9 & 97.8 \\

FlashVID~\cite{fan2026flashvid} \textcolor{gray}{(ICLR'26)}
&
& 74.0 & 60.0 & 51.9 & 62.0
& \underline{58.2} & \underline{56.7}
& \underline{58.3} & 59.5
& \underline{59.1} & \underline{98.1} \\

\textbf{MTRC (Ours)}
&
& \ourscell{\textbf{74.2}}
& \ourscell{\underline{60.4}}
& \ourscell{\textbf{52.3}}
& \ourscell{\textbf{62.3}}
& \ourscell{57.4}
& \ourscell{\textbf{56.8}}
& \ourscell{\textbf{58.7}}
& \ourscell{\textbf{60.5}}
& \ourscell{\textbf{59.6}}
& \ourscell{\textbf{98.9}} \\

\midrule


VisionZip~\cite{yang2025visionzip} \textcolor{gray}{(CVPR'25)}
& \multirow{5}{*}{15\%}
& 72.9 & 58.1 & \textbf{51.9} & 61.0
& \textbf{58.2} & \textbf{55.7}
& 57.5 & 59.1 & 58.3 & 96.8 \\

FastVID~\cite{shen2026fastvid} \textcolor{gray}{(NeurIPS'25)}
&
& 73.4 & 58.1 & \underline{51.8} & 61.1
& 56.8 & 54.2
& 57.7 & \underline{60.1} & 58.3 & 96.7 \\

FlashVID~\cite{fan2026flashvid} \textcolor{gray}{(ICLR'26)}
&
& \underline{73.6} & \underline{58.9}
& 51.5 & \underline{61.3}
& \underline{57.2} & \underline{55.4}
& \underline{58.0} & 60.0
& \underline{58.7} & \underline{97.4} \\

ApET~\cite{ma2026apet} \textcolor{gray}{(CVPR'26)}
&
& 72.7 & 58.1 & 51.0 & 60.6
& 53.8 & 54.1
& 57.2 & \underline{60.1} & 58.0 & 96.3 \\

\textbf{MTRC (Ours)}
&
& \ourscell{\textbf{74.0}}
& \ourscell{\textbf{59.6}}
& \ourscell{\underline{51.8}}
& \ourscell{\textbf{61.8}}
& \ourscell{57.0}
& \ourscell{55.1}
& \ourscell{\textbf{58.7}}
& \ourscell{\textbf{60.2}}
& \ourscell{\textbf{59.0}}
& \ourscell{\textbf{97.8}} \\

\midrule


VisionZip~\cite{yang2025visionzip} \textcolor{gray}{(CVPR'25)}
& \multirow{4}{*}{10\%}
& 69.2 & \textbf{57.9} & \textbf{51.2} & 59.4
& 54.4 & \textbf{53.9}
& 54.8 & 58.4 & 56.6 & 94.0 \\

FastVID~\cite{shen2026fastvid} \textcolor{gray}{(NeurIPS'25)}
&
& 71.6 & 57.3 & 50.1 & \underline{59.7}
& \underline{54.8} & 52.4
& \textbf{56.9} & \underline{59.3}
& 57.1 & 94.7 \\

FlashVID~\cite{fan2026flashvid} \textcolor{gray}{(ICLR'26)}
&
& \underline{71.8} & 57.1 & 50.2 & \underline{59.7}
& \textbf{55.2} & \textbf{53.9}
& \underline{56.8} & 59.1
& \underline{57.4} & \underline{95.2} \\

\textbf{MTRC (Ours)}
&
& \ourscell{\textbf{71.9}}
& \ourscell{\underline{57.8}}
& \ourscell{\underline{50.8}}
& \ourscell{\textbf{60.1}}
& \ourscell{\underline{54.8}}
& \ourscell{\underline{53.6}}
& \ourscell{\textbf{56.9}}
& \ourscell{\textbf{59.6}}
& \ourscell{\textbf{57.6}}
& \ourscell{\textbf{95.5}} \\

\bottomrule
\end{tabular}%
}

\endgroup

\smallskip

\end{table*}

\section{Experiments}
\subsection{Experimental Setup}
\noindent\textbf{Benchmarks.}
We evaluate MTRC on four benchmarks: Video-MME~\cite{fu2025video}, EgoSchema~\cite{mangalam2023egoschema}, LongVideoBench~\cite{wu2024longvideobench}, and MVBench~\cite{li2024mvbench}. These benchmarks cover different video durations and diverse video understanding tasks, enabling a comprehensive evaluation of our method's effectiveness and generalization across different scenarios.

\noindent\textbf{Compared baselines.}
We evaluate our method on two representative VLLMs, LLaVA-OV-7B~\cite{li2025llavaonevisioneasyvisualtasktransfer} and LLaVA-Video~\cite{zhang2025llavavideo}, to examine the transferability of MTRC across model backbones. We compare MTRC with five recent representative training-free VLLM acceleration methods: VisionZip~\cite{yang2025visionzip}, PruneVID~\cite{huang2025prunevid}, FastVID~\cite{shen2026fastvid}, FlashVID~\cite{fan2026flashvid}, and EarlyTom~\cite{wang2026earlytom}. To compare methods at the same visual token retention ratio, we disable their inner-LLM pruning and retain only token compression performed before entering the LLM.

\noindent\textbf{Implementation details.}
We uniformly sample 32 frames for LLaVA-OneVision and
64 for LLaVA-Video, yielding $32\times196$ and
$64\times169$ visual patch tokens, respectively.
We evaluate token retention ratios of
10\%, 15\%, 20\%, and 25\%.
Default hyperparameters are shared across settings,
except for $S$ and $\tau$.
Accuracy and efficiency are evaluated on NVIDIA
RTX 4090 and A100 GPUs, respectively.
FlashAttention-2~\cite{dao2024flashattention} is enabled
for all compatible methods.
Additional implementation details are provided in
Appendix~\ref{app:implementation_details}.

\subsection{Main Results}

We compare MTRC with state-of-the-art baselines across different retention ratios on two representative VLLMs.

\noindent\textbf{Results on LLaVA-OneVision.}
Table~\ref{tab:strc_main} compares MTRC with existing training-free token
compression methods on four video understanding benchmarks.
MTRC achieves the highest average score among the compared
methods at all four retention ratios: 25\%, 20\%, 15\%, and 10\%.
Notably, MTRC achieves 100.2\% of the uncompressed model's
average performance while retaining only 15\% of visual
patch tokens.
When the retention ratio is further reduced to 10\%,
MTRC still preserves 98.1\% of the uncompressed model's
average performance.
These results demonstrate that MTRC maintains strong video
understanding performance even under low token budgets.

\noindent\textbf{Results on LLaVA-Video.}
As shown in Table~\ref{tab:llava_video_main_provisional}, MTRC achieves the highest average score at all retention ratios, although some baselines
remain competitive on individual benchmarks.
Together with the results on LLaVA-OneVision, these findings
demonstrate the effectiveness of the proposed spatiotemporal
reconstruction-based compression strategy across both model
backbones, supporting its cross-model applicability.

In addition, we evaluate whether MTRC can improve video understanding
by processing more frames under a fixed visual token budget.
We also combine MTRC with FastV-style decoder pruning to examine
whether cross-stage compression provides complementary accuracy gains.
Details and results are provided in Appendix~\ref{sec:supp_extensions}.

\begin{table*}[t]
\centering
\caption{Efficiency and accuracy summary on LLaVA-OneVision-7B.
Latencies are reported in milliseconds.}
\label{tab:efficiency_accuracy}

\begingroup
\small
\setlength{\tabcolsep}{9.5pt}
\renewcommand{\arraystretch}{1.0}

\resizebox{\textwidth}{!}{%
\begin{tabular}{@{}lccccccc@{}}
\toprule

\multirow{2}{*}{Method}
& \multirow{2}{*}{\makecell{Retention\\$R$}}
& \multirow{2}{*}{TFLOPs $\downarrow$}
& \multirow{2}{*}{\makecell{Vision\\Encoding}}
& \multirow{2}{*}{\makecell{LLM Prefilling\\$\downarrow$}}
& \multirow{2}{*}{TTFT $\downarrow$}
& \multicolumn{2}{c}{Avg.} \\

\cmidrule(lr){7-8}
& & & & & & Score $\uparrow$ & Rel.\ Acc.\ (\%) $\uparrow$ \\

\midrule

LLaVA-OneVision
& 100\% & 99.62 & 374.45
& 566.79 ($1.0\times$)
& 1213.50 ($1.0\times$)
& 58.4 & 100.0 \\

\midrule

VisionZip~\cite{yang2025visionzip} \textcolor{gray}{(CVPR'25)}
& 25\% & 23.48 & 376.59
& 163.35 ($3.5\times$)
& \underline{643.23} ($1.9\times$)
& \underline{58.1} & \underline{99.5} \\

FastVID~\cite{shen2026fastvid} \textcolor{gray}{(NeurIPS'25)}
& 25\% & 23.01 & 374.96
& 165.26 ($3.4\times$)
& 665.36 ($1.8\times$)
& 57.1 & 97.8 \\

EarlyTom~\cite{wang2026earlytom} \textcolor{gray}{(CVPR'26)}
& 25\% & 23.09 & \textbf{337.19}
& 165.48 ($3.4\times$)
& 649.14 ($1.9\times$)
& \underline{58.1} & 99.4 \\

FlashVID~\cite{fan2026flashvid} \textcolor{gray}{(ICLR'26)}
& 15\% & \underline{14.59} & \underline{372.71}
& \underline{101.73} ($5.6\times$)
& 677.20 ($1.8\times$)
& 58.0 & 99.3 \\

\midrule

\textbf{MTRC (Ours)}
& 15\%
& \ourscell{\textbf{14.15}}
& \ourscell{375.71}
& \ourscell{\textbf{99.38} ($5.7\times$)}
& \ourscell{\textbf{638.24} ($1.9\times$)}
& \ourscell{\textbf{58.5}}
& \ourscell{\textbf{100.2}} \\

\bottomrule
\end{tabular}%
}

\endgroup
\end{table*}

\subsection{Efficiency Analysis}
Table~\ref{tab:efficiency_accuracy} summarizes the accuracy and inference latency of LLaVA-OneVision-7B evaluated with 32 sampled frames on a single NVIDIA A100 GPU.
With 15\% of visual patch tokens retained, MTRC preserves
100.2\% of the uncompressed model's average score across
four benchmarks, while achieving a $5.7\times$ speedup
in pure LLM prefilling and a $1.9\times$ speedup in
time to first token (TTFT) relative to the uncompressed model.
At the same 15\% retention ratio, MTRC reduces TTFT
by approximately 5.8\% compared with FlashVID.
Furthermore, compared with VisionZip, FastVID, and EarlyTom
at 25\% retention, MTRC achieves a higher average score
at the lower retention ratio of 15\%, with slightly lower
reported TTFT than all three methods.
These results demonstrate that MTRC maintains competitive
accuracy and inference latency with fewer visual tokens.

\begin{figure*}[t]
    \centering
 \includegraphics[width=0.97\textwidth]{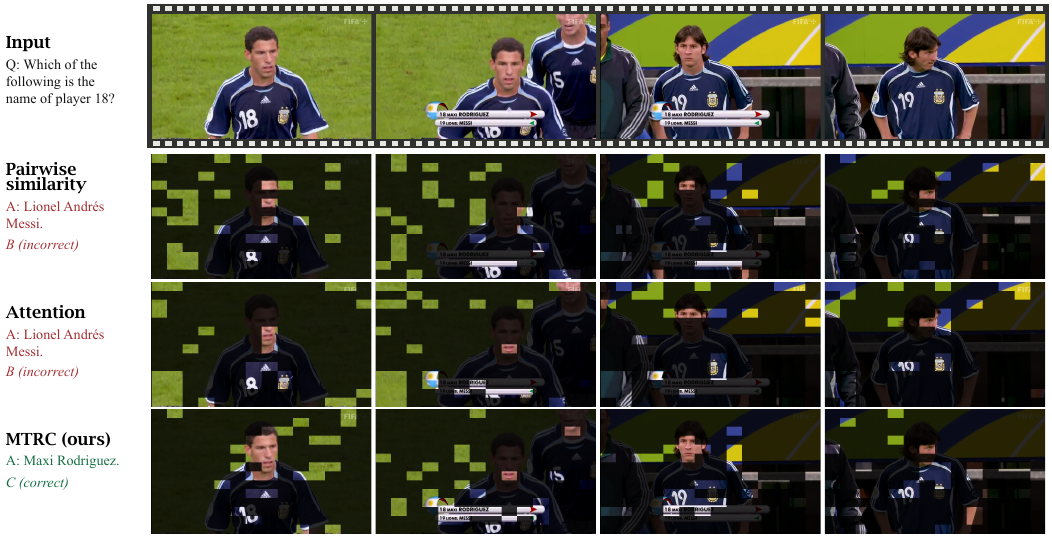}
    \caption{Qualitative comparison of MTRC with attention-based and
pairwise-similarity-based baselines at 15\% visual token retention.
Retained tokens are highlighted.} \label{fig:qualitative}
\end{figure*}
\subsection{Ablation Studies}
Unless otherwise specified, we conduct ablation studies on
Video-MME and LongVideoBench using LLaVA-OneVision-7B with
15\% visual patch token retention.

\begin{table}[t]
\centering
\caption{
Ablation of temporal selection and spatial filtering.
}
\label{tab:module_ablation}

\small
\setlength{\tabcolsep}{3pt}
\renewcommand{\arraystretch}{1.1}

\resizebox{\columnwidth}{!}{%
\begin{tabular}{@{}cccccccc@{}}
\toprule

\multirow{2}{*}{TSM}
& \multirow{2}{*}{SFM}
& \multicolumn{4}{c}{VideoMME}
& \multirow{2}{*}[-2pt]{\shortstack{LongVideo\\Bench}}
& \multirow{2}{*}[-2pt]{\shortstack{Rel. Acc.\\(\%)}} \\

\cmidrule(lr){3-6}

 &  & Short & Medium & Long & Overall & & \\

\midrule

$\checkmark$ & $\times$
& 68.8 & 54.6 & 49.2 & 57.5 & 55.9 & 98.6 \\

$\times$ & $\checkmark$
& 69.1 & 53.8 & 49.1 & 57.3 & 56.1 & 98.6 \\

$\checkmark$ & $\checkmark$
& \ourscell{\textbf{69.9}}
& \ourscell{\textbf{55.2}}
& \ourscell{\textbf{50.3}}
& \ourscell{\textbf{58.5}}
& \ourscell{\textbf{57.5}}
& \ourscell{\textbf{100.9}} \\

\bottomrule
\end{tabular}%
}
\end{table}

\noindent\textbf{Complementarity of TSM and SFM.}
As shown in Table~\ref{tab:module_ablation}, we compare three configurations: TSM only, SFM only, and their combination.
TSM reduces the repeated retention of existing information across frames, while SFM further reduces within-frame redundancy in the candidate pool.
Under the same token retention budget, combining the two modules achieves better video understanding performance, indicating that temporal selection and spatial filtering are complementary.

\noindent\textbf{Sensitivity to the temporal threshold.}
 As shown in Table~\ref{tab:tau_ablation}, the best-performing tested thresholds differ between the two backbones, indicating that a single threshold may not suit different models. We therefore use temporal thresholds of 0.57 and 0.60 for LLaVA-OneVision and LLaVA-Video, respectively. The values 0.57 and 0.55 correspond to the empirical medians of temporal reconstruction errors for LLaVA-OneVision and LLaVA-Video, respectively.
\begin{table}[t]
\centering
\caption{
Sensitivity to the temporal threshold $\tau$.
}
\label{tab:tau_ablation}

\footnotesize
\begin{tabular}{@{}ccccccc@{}}
\toprule
\multirow[c]{2}{*}[-2pt]{$\tau$}
& \multicolumn{4}{c}{VideoMME}
& \multirow[c]{2}{*}[-2pt]{\makecell[c]{LongVideo\\Bench}}
& \multirow[c]{2}{*}[-2pt]{\makecell[c]{Rel.\ Acc.\\(\%)}} \\
\cmidrule(lr){2-5}
& Short & Medium & Long & Overall & & \\
\midrule

\multicolumn{7}{c}{\textbf{LLaVA-Video-7B (64 frames)}} \\
\midrule

0.50
& \textbf{74.2} & 59.0 & 50.6 & 61.3 & 58.1 & 97.3 \\

0.55
& 73.3 & \textbf{59.7} & 50.6 & 61.2 & 57.7 & 96.9 \\

0.60
& 74.0 & 59.6 & \textbf{51.8}
& \textbf{61.8} & \textbf{58.7} & \textbf{98.2} \\

\midrule
\multicolumn{7}{c}{\textbf{LLaVA-OneVision-7B (32 frames)}} \\
\midrule

0.50
& \textbf{70.2} & 54.0 & 49.8 & 58.0 & 56.4 & 99.5 \\

0.57
& 69.9 & \textbf{55.2} & \textbf{50.3}
& \textbf{58.5} & \textbf{57.5} & \textbf{100.9} \\

0.60
& 69.0 & 54.8 & 49.2 & 57.7 & 56.1 & 98.9 \\

\bottomrule
\end{tabular}
\end{table}
\begin{table}[t]
\centering
\caption{
Effects of the retrieved historical token count $B$ and
history window size $H$.
$B=1^\dagger$ matches the number of candidates passed to SFM
under $B=10$ in each frame.
}
\label{tab:history_ablation}

\footnotesize
\begin{tabular}{@{}ccccc@{}}
\toprule
\makecell[c]{Parameter}
& \makecell[c]{Value}
& \makecell[c]{VideoMME}
& \makecell[c]{LongVideo\\Bench}
& \makecell[c]{Rel.\ Acc.\\(\%)} \\
\midrule

\multirow{5}{*}{$B$}
& 1 & 57.4 & 56.5 & 99.0 \\
& $1^\dagger$ & 57.1 & 56.2 & 98.5 \\
& 5 & 58.1 & 56.9 & 100.0 \\
& \textbf{10} & \textbf{58.5}
& \textbf{57.5} & \textbf{100.9} \\
& 15 & 57.6 & 56.8 & 99.5 \\

\midrule

\multirow{3}{*}{$H$}
& 2 & 58.1 & 56.7 & 99.8 \\
& \textbf{4} & \textbf{58.5}
& \textbf{57.5} & \textbf{100.9} \\
& 6 & 58.2 & 56.7 & 99.9 \\

\bottomrule
\end{tabular}
\end{table}

\noindent\textbf{Effects of historical token count and window size.}
As shown in Table~\ref{tab:history_ablation}, $B=10$ and $H=4$ achieved the best performance among the settings tested in their respective ablations and were therefore selected as defaults.
To control for differences in candidate count, we additionally introduce a $B=1^\dagger$ control.
For each frame, this control matches the number of candidate tokens passed to SFM under $B=10$, with the rest of the pipeline unchanged.
Compared with this control, $B=10$ still achieves higher accuracy, indicating that the gains from multi-token temporal reconstruction are not solely attributable to changes in candidate count.

\noindent\textbf{Effect of multi-token reconstruction.}
We replace reconstruction-based
scoring in both TSM and SFM with pairwise similarity to the most
similar reference token.
The reference tokens come from historical frames in TSM and the
current frame in SFM.
We match the number of candidates passed to SFM frame by frame
while keeping the rest of the pipeline unchanged.
As shown in Table~\ref{tab:pairwise_vs_reconstruction},
MTRC achieves higher accuracy than the pairwise-similarity baseline
on both benchmarks.
These results support the effectiveness of multi-token reconstruction
for spatiotemporal redundancy assessment.

\begin{table}[t]
\centering
\caption{
Comparison of pairwise similarity and multi-token reconstruction
for spatiotemporal redundancy assessment.
}
\label{tab:pairwise_vs_reconstruction}

\resizebox{\columnwidth}{!}{%
\begin{tabular}{@{}lcccc@{}}
\toprule
Scoring method
& \makecell{VideoMME}
& \makecell{LongVideo\\Bench}
& \makecell{Avg.\\Score}
& \makecell{Rel.\ Acc.\\(\%)} \\
\midrule

Pairwise similarity
& 56.7 & 56.3 & 56.5
& 98.3 \\

\textbf{MTRC (Ours)}
& \ourscell{\textbf{58.5}}
& \ourscell{\textbf{57.5}}
& \ourscell{\textbf{58.0}}
& \ourscell{\textbf{100.9}} \\

\bottomrule
\end{tabular}%
}
\end{table}

\noindent\textbf{Visualization.}
Figure~\ref{fig:qualitative} compares MTRC with attention-based
and pairwise-similarity-based baselines.
The attention baseline ranks tokens by attention scores,
while the similarity baseline separately evaluates maximum cosine
similarity to retained historical tokens and same-frame basis tokens.
Compared with the two baselines, MTRC better preserves key visual cues,
including players' appearances, jersey numbers, and on-screen captions.
These cues provide complementary evidence for associating the player
wearing number~18 with the name displayed in the captions.
In this example, MTRC answers the question correctly,
whereas both baselines produce an incorrect answer.

Additional ablation studies on temporal reconstruction weights,
spatial basis size, and reconstruction weight constraints are provided
in Appendix~\ref{sec:supp_ablation}.
Further qualitative examples comparing the token selection methods
are presented in Appendix~\ref{sec:supp_qualitative}.
\section{Conclusion}

In this paper, we rethink video token compression from the perspective of spatiotemporal basis selection and propose MTRC, a training-free framework for video large language models.
The framework uses reconstruction errors to assess the joint representational capacity of multiple tokens.
Through the complementary integration of the Temporal Selection Module (TSM) and the Spatial Filtering Module (SFM), it reduces cross-frame and within-frame redundancy, constructing a compact spatiotemporal basis set under a fixed budget.
Notably, on LLaVA-OneVision, MTRC retains only 15\% of visual tokens while maintaining accuracy comparable to the uncompressed baseline and achieving a $1.90\times$ speedup in time to first token (TTFT).
These results indicate that spatiotemporal basis selection provides an effective approach to balancing video understanding performance and inference efficiency.


{
    \small
    \bibliographystyle{ieeenat_fullname}
    \bibliography{main}
}

\clearpage
\maketitlesupplementary
\appendix
\makeatletter
\setlength{\@fptop}{0pt}
\setlength{\@dblfptop}{0pt}
\makeatother

\setcounter{table}{0}
\renewcommand{\thetable}{S\arabic{table}}
\setcounter{figure}{0}
\renewcommand{\thefigure}{S\arabic{figure}}
\providecommand{\ourscell}[1]{\cellcolor[gray]{0.92}#1}

\section*{Overview}

In this supplement, we present extended evaluations, additional ablation
studies, implementation details, and qualitative visualizations of MTRC.
It is organized as follows:

\begin{itemize}
   \item \textbf{Section~\ref{sec:supp_extensions}} examines whether MTRC
can benefit from more input frames under a fixed visual token budget.
We compare MTRC, VisionZip, and FastVID on Video-MME and LongVideoBench
using LLaVA-OneVision-7B with 128 or 256 input frames per video.
Both frame counts use the same budget as the 32-frame baseline:
6,272 visual patch tokens per video.
We also compare MTRC with and without FastV-style decoder pruning
on LLaVA-OneVision-7B. Both pruning settings use 32 input frames
per video and retain approximately 15\% of visual patch tokens
on average across decoder layers.

   \item \textbf{Section~\ref{sec:supp_ablation}} presents additional
ablation studies on Video-MME and LongVideoBench using LLaVA-OneVision-7B.
We compare three weighting schemes for temporal reconstruction:
MTRC's optimized weights, uniform averaging, and similarity weighting.
At each visual token retention ratio, we fix the number of spatial basis
tokens at 10 for the first frame and vary it for subsequent frames.
Finally, we examine the effects of imposing nonnegativity and sum-to-one
constraints on the temporal reconstruction weights, both individually
and jointly.

    \item \textbf{Section~\ref{app:implementation_details}} provides
additional implementation details, including evaluation settings,
reconstruction hyperparameters, and the candidate-pool construction rule.

     \item 
     \textbf{Section~\ref{sec:supp_qualitative}} presents qualitative
comparisons of MTRC with pairwise-similarity-based and attention-based
baselines, showing retained visual tokens and the corresponding
model answers.
\end{itemize}

\section{Extended Evaluations}
\label{sec:supp_extensions}

We conduct extended evaluations on Video-MME and LongVideoBench
using LLaVA-OneVision-7B as the backbone.
These evaluations examine MTRC's performance with more input frames
under a fixed visual token budget and its compatibility with
decoder-side visual token pruning.

\par\addvspace{1.5\baselineskip}
\refstepcounter{subsection}
\noindent\textbf{More Input Frames under a Fixed Token Budget.}
Due to computational and memory constraints, existing video large language models typically process relatively few sampled frames, often missing important visual cues.
To evaluate MTRC's ability to utilize more video frames, we increase the input to 128 and 256 frames under a fixed budget of 6,272 visual patch tokens.
As shown in Table~\ref{tab:more_frames}, MTRC outperforms both the uncompressed 32-frame baseline and the compared methods in both settings, demonstrating its ability to improve video understanding with more frames under a fixed token budget.

\begin{table}[!t]
\centering
\caption{Performance comparison with increased input frames.}
\label{tab:more_frames}

\begingroup
\footnotesize
\setlength{\tabcolsep}{0pt}
\providecommand{\ourscell}[1]{\cellcolor[gray]{0.92}#1}

\begin{tabular*}{\columnwidth}{@{}
  >{\raggedright\arraybackslash}m{.218\columnwidth}
  >{\centering\arraybackslash}m{.142\columnwidth}
  >{\centering\arraybackslash}m{.135\columnwidth}
  >{\centering\arraybackslash}m{.102\columnwidth}
  >{\centering\arraybackslash}m{.157\columnwidth}
  >{\centering\arraybackslash}m{.110\columnwidth}
  >{\centering\arraybackslash}m{.136\columnwidth}
@{}}
\toprule
Method
& Frames
& \shortstack{Retention\\$R$}
& \shortstack{Video\\MME}
& \shortstack{LongVideo\\Bench}
& \shortstack{Avg.\\Score}
& \shortstack{Rel.\ Acc.\\(\%)} \\
\midrule

LLaVA-OV-7B
& 32 & 100\%
& 58.4 & 56.6 & 57.5 & 100.0 \\
\midrule

\makecell[l]{VisionZip~\cite{yang2025visionzip}\\
             \textcolor{gray}{(CVPR'25)}}
& \multirow{3}{*}[-6pt]{128 ($4\times$)}
& \multirow{3}{*}[-6pt]{25\%}
& 59.6 & 57.9 & 58.8 & 102.2 \\

\makecell[l]{FastVID~\cite{shen2026fastvid}\\
             \textcolor{gray}{(NeurIPS'25)}}
& &
& 57.8 & 56.1 & 57.0 & 99.0 \\

\textbf{MTRC (Ours)}
& &
& \ourscell{\textbf{60.5}}
& \ourscell{\textbf{58.2}}
& \ourscell{\textbf{59.4}}
& \ourscell{\textbf{103.2}} \\
\midrule

\makecell[l]{VisionZip~\cite{yang2025visionzip}\\
             \textcolor{gray}{(CVPR'25)}}
& \multirow{3}{*}[-6pt]{256 ($8\times$)}
& \multirow{3}{*}[-6pt]{12.5\%}
& 58.5 & 54.8 & 56.7 & 98.5 \\

\makecell[l]{FastVID~\cite{shen2026fastvid}\\
             \textcolor{gray}{(NeurIPS'25)}}
& &
& 56.1 & 53.4 & 54.8 & 95.2 \\

\textbf{MTRC (Ours)}
& &
& \ourscell{\textbf{60.4}}
& \ourscell{\textbf{57.9}}
& \ourscell{\textbf{59.2}}
& \ourscell{\textbf{102.9}} \\
\bottomrule
\end{tabular*}
\endgroup
\end{table}

\par\addvspace{1.5\baselineskip}
\refstepcounter{subsection}
\noindent\textbf{Extended Experiments on Cross-Stage Compression.}
We combine MTRC with FastV-style decoder pruning and compare it with PruneVid and FlashVID at a layer-averaged retention ratio of 15\%.
As shown in Table~\ref{tab:supp_mtrc_fastv}, without additional hyperparameter tuning for the combination, MTRC + FastV outperforms both MTRC alone and the two competing methods on both benchmarks.
These results demonstrate that MTRC can be directly combined with decoder-side pruning to further improve video understanding performance.

\begin{table}[!t]
\centering
\caption{Extended experiments on cross-stage compression. $R$ denotes the visual patch token retention ratio averaged
over the 28 decoder layers.}
\label{tab:supp_mtrc_fastv}

\begingroup
\footnotesize
\setlength{\tabcolsep}{.004\columnwidth}
\providecommand{\ourscell}[1]{\cellcolor[gray]{0.92}#1}

\begin{tabular}{
  >{\centering\arraybackslash}m{.281\columnwidth}
  >{\centering\arraybackslash}m{.149\columnwidth}
  >{\centering\arraybackslash}m{.165\columnwidth}
  >{\centering\arraybackslash}m{.157\columnwidth}
  >{\centering\arraybackslash}m{.079\columnwidth}
  >{\centering\arraybackslash}m{.121\columnwidth}
}
\toprule
Method
& \shortstack{Retention\\$R$}
& VideoMME
& \shortstack{LongVideo\\Bench}
& \shortstack{Avg.\\Score}
& \shortstack{Rel.\,Acc.\\(\%)} \\
\midrule


PruneVid~\cite{huang2025prunevid}
& & & & & \\

\textcolor{gray}{(ACL Findings'25)}
& \multirow[c]{-2}{*}{15\%}
& \multirow[c]{-2}{*}{56.4}
& \multirow[c]{-2}{*}{54.6}
& \multirow[c]{-2}{*}{55.5}
& \multirow[c]{-2}{*}{96.5} \\

\addlinespace[2pt]

FlashVID~\cite{fan2026flashvid}
& & & & & \\

\textcolor{gray}{(ICLR'26)}
& \multirow[c]{-2}{*}{15\%}
& \multirow[c]{-2}{*}{58.0}
& \multirow[c]{-2}{*}{57.2}
& \multirow[c]{-2}{*}{57.6}
& \multirow[c]{-2}{*}{100.2} \\

\midrule

MTRC
& 15\%
& 58.5
& 57.5
& 58.0
& 100.9 \\

\textbf{MTRC + FastV}~\cite{chen2024image}
&
& \ourscell{}
& \ourscell{}
& \ourscell{}
& \ourscell{} \\

\textcolor{gray}{(ECCV'24)}
& \multirow[c]{-2}{*}{15\%}
& \ourscell{\multirow[c]{-2}{*}{\textbf{58.8}}}
& \ourscell{\multirow[c]{-2}{*}{\textbf{57.9}}}
& \ourscell{\multirow[c]{-2}{*}{\textbf{58.4}}}
& \ourscell{\multirow[c]{-2}{*}{\textbf{101.5}}} \\

\bottomrule
\end{tabular}
\endgroup
\end{table}

\section{Additional Ablation Studies}
\label{sec:supp_ablation}
In this section, we conduct additional ablation studies on
Video-MME and LongVideoBench using LLaVA-OneVision-7B as the backbone.
We examine how temporal reconstruction weights, spatial basis size,
and reconstruction weight constraints affect video understanding
performance.
In addition, we examine how temporal reconstruction weights, spatial basis size, and reconstruction weight constraints affect model performance.

\par\addvspace{1.5\baselineskip}
\refstepcounter{subsection}
\noindent\textbf{Temporal Reconstruction Weights.}
To examine the effect of temporal reconstruction weights in TSM, we replace MTRC's optimized reconstruction weights with uniform or similarity-based weights, while keeping the rest of the compression pipeline unchanged.
MTRC obtains its optimized weights through Frank--Wolfe iterations under the convex-combination constraint.
As shown in Table~\ref{tab:supp_trf_weights}, the optimized weighting strategy outperforms both alternatives on both benchmarks.
These results suggest that optimizing combination weights to minimize joint reconstruction error can improve video understanding performance compared with uniform or purely similarity-based weighting.

\begin{table}[!t]
\centering
\caption{Ablation of temporal reconstruction weights in TSM.}
\label{tab:supp_trf_weights}

\begingroup
\footnotesize
\setlength{\tabcolsep}{0pt}
\providecommand{\ourscell}[1]{\cellcolor[gray]{0.92}#1}

\begin{tabular*}{\columnwidth}{@{}
  >{\raggedright\arraybackslash}m{.34\columnwidth}
  >{\centering\arraybackslash}m{.17\columnwidth}
  >{\centering\arraybackslash}m{.19\columnwidth}
  >{\centering\arraybackslash}m{.13\columnwidth}
  >{\centering\arraybackslash}m{.17\columnwidth}
@{}}
\toprule
Weighting rule
& \shortstack{VideoMME}
& \shortstack{LongVideo\\Bench}
& \shortstack{Avg.\\Score}
& \shortstack{Rel.\ Acc.\\(\%)} \\
\midrule

Uniform averaging
& 57.3 & 56.3 & 56.8 & 98.8 \\

Similarity weighting
& 57.5 & 56.5 & 57.0 & 99.1 \\

\textbf{Frank--Wolfe (Ours)}
& \ourscell{\textbf{58.5}}
& \ourscell{\textbf{57.5}}
& \ourscell{\textbf{58.0}}
& \ourscell{\textbf{100.9}} \\
\bottomrule
\end{tabular*}

\endgroup
\end{table}

\par\addvspace{1.5\baselineskip}
\refstepcounter{subsection}
\noindent\textbf{Effect of the Spatial Basis Size.}
We fix the first-frame spatial basis size at 10 and vary only $S$ for subsequent frames.
Under a fixed budget, too many spatial basis tokens reduce the budget available for representative tokens, limiting the retention of complementary information.
This effect is particularly pronounced at the low retention ratio of 10\%.
As shown in Table~\ref{tab:supp_spatial_basis_size}, $S=4$ achieves the highest average accuracy at 10\% retention, whereas $S=10$ performs best at the other retention ratios.
We therefore adopt these configurations as the default settings.

\begin{table}[!t]
\centering
\caption{Ablation of the spatial basis size $S$.
Shaded rows indicate the default settings:
$S=4$ at 10\% retention and $S=10$ otherwise.}
\label{tab:supp_spatial_basis_size}

\begingroup
\footnotesize
\setlength{\tabcolsep}{0pt}
\providecommand{\ourscell}[1]{\cellcolor[gray]{0.92}#1}

\begin{tabular*}{\columnwidth}{@{}
  >{\centering\arraybackslash}m{.16\columnwidth}
  >{\centering\arraybackslash}m{.14\columnwidth}
  >{\centering\arraybackslash}m{.20\columnwidth}
  >{\centering\arraybackslash}m{.20\columnwidth}
  >{\centering\arraybackslash}m{.14\columnwidth}
  >{\centering\arraybackslash}m{.16\columnwidth}
@{}}
\toprule
\shortstack{Retention\\$R$}
& \shortstack{Basis size\\$S$}
& VideoMME
& \shortstack{LongVideo\\Bench}
& \shortstack{Avg.\\Score}
& \shortstack{Rel.\ Acc.\\(\%)} \\
\midrule


\multirow{4}{*}{25\%}
& 6
& 58.6 & 56.8 & 57.7 & 100.3 \\

& 8
& 58.4 & 56.9 & 57.7 & 100.3 \\

& \ourscell{\textbf{10}}
& \ourscell{\textbf{58.7}}
& \ourscell{\textbf{57.4}}
& \ourscell{\textbf{58.1}}
& \ourscell{\textbf{101.0}} \\

& 12
& 58.6 & 56.5 & 57.6 & 100.1 \\
\midrule

\multirow{4}{*}{20\%}
& 6
& 58.1 & 56.2 & 57.2 & 99.4 \\

& 8
& 58.3 & 56.3 & 57.3 & 99.7 \\

& \ourscell{\textbf{10}}
& \ourscell{\textbf{58.1}}
& \ourscell{\textbf{56.9}}
& \ourscell{\textbf{57.5}}
& \ourscell{\textbf{100.0}} \\

& 12
& 58.3 & 56.1 & 57.2 & 99.5 \\
\midrule

\multirow{4}{*}{15\%}
& 6
& 58.1 & 57.1 & 57.6 & 100.2 \\

& 8
& 57.9 & 57.4 & 57.7 & 100.3 \\

& \ourscell{\textbf{10}}
& \ourscell{\textbf{58.5}}
& \ourscell{\textbf{57.5}}
& \ourscell{\textbf{58.0}}
& \ourscell{\textbf{100.9}} \\

& 12
& 58.1 & 57.6 & 57.9 & 100.6 \\
\midrule

\multirow{4}{*}{10\%}
& \ourscell{\textbf{4}}
& \ourscell{\textbf{57.1}}
& \ourscell{\textbf{55.6}}
& \ourscell{\textbf{56.4}}
& \ourscell{\textbf{98.0}} \\

& 6
& 56.3 & 55.2 & 55.8 & 97.0 \\

& 8
& 56.3 & 55.2 & 55.8 & 97.0 \\

& 10
& 56.1 & 54.8 & 55.5 & 96.4 \\

\bottomrule
\end{tabular*}
\endgroup
\end{table}

\par\addvspace{1.5\baselineskip}
\refstepcounter{subsection}
\noindent\textbf{Effect of Reconstruction Weight Constraints.}
To examine the role of convex-combination constraints, we compare four
temporal reconstruction settings: no constraints, nonnegativity alone,
a sum-to-one constraint alone, and both constraints together.
As shown in Table~\ref{tab:reconstruction_constraints}, imposing both
constraints yields better video understanding performance than imposing
either constraint alone or neither constraint.

\begin{table}[t]
\centering
\caption{
Ablation of constraints on reconstruction weights.
$\checkmark$ indicates that the corresponding constraint is imposed.}
\label{tab:reconstruction_constraints}

\begingroup
\footnotesize
\setlength{\tabcolsep}{3pt}
\renewcommand{\arraystretch}{1.1}

\resizebox{\columnwidth}{!}{%
\begin{tabular}{@{}lcccccc@{}}
\toprule

\makecell[c]{Reconstruction\\method}
& \makecell[c]{Nonnegative\\$\alpha \geq 0$}
& \makecell[c]{Sum-to-one\\$\mathbf{1}^{\top}\alpha = 1$}
& \makecell[c]{VideoMME}
& \makecell[c]{LongVideo\\Bench}
& \makecell[c]{Avg.\\Score}
& \makecell[c]{Rel.\ Acc.\\(\%)} \\

\midrule

Unconstrained
& $\times$ & $\times$
& 57.9 & 57.2 & 57.6 & 100.1 \\

Nonnegative
& $\checkmark$ & $\times$
& 57.8 & 56.3 & 57.1 & 99.2 \\

Affine
& $\times$ & $\checkmark$
& 58.2 & 56.6 & 57.4 & 99.8 \\

\textbf{Convex (Ours)}
& $\checkmark$ & $\checkmark$
& \ourscell{\textbf{58.5}}
& \ourscell{\textbf{57.5}}
& \ourscell{\textbf{58.0}}
& \ourscell{\textbf{100.9}} \\

\bottomrule
\end{tabular}%
}

\endgroup
\end{table}

\section{Additional Implementation Details}
\label{app:implementation_details}

In this section, we provide additional implementation details to facilitate the reproduction of MTRC.
We specify the evaluation settings and the hyperparameters used for temporal selection and spatial filtering.
We also describe how the candidate pool is constructed and supplemented when necessary to provide sufficient candidates for subsequent spatial filtering.

\par\addvspace{1.5\baselineskip}
\noindent\textbf{Evaluation settings.}
For LLaVA-Video, we follow FlashVID and adopt
the frame-level newline-token setting.
All efficiency comparisons use the same hardware
platform and a unified testing protocol.

\par\addvspace{1.5\baselineskip}
\noindent\textbf{Hyperparameter settings.}
We use model accuracy to select the number of retrieved historical
tokens $B$, the history window size $H$, the temporal filtering
threshold $\tau$, and the number of spatial basis tokens $S$ for
frames after the first.
Unless otherwise specified, we use a history window of $H=4$ frames
at all retention ratios. For each current token, we retrieve up to
$B=10$ tokens from the historical memory to form its temporal dictionary.
We set $\tau=0.57$ for LLaVA-OneVision and $\tau=0.60$ for LLaVA-Video.
For the first frame, we use DPC-KNN with $k=5$ neighbors to select
a fixed set of $S=10$ spatial basis tokens.
For subsequent frames, we use $S=4$ at 10\% retention and $S=10$
at the other retention ratios, based on Table~\ref{tab:supp_spatial_basis_size}.
We use $Q=4$ Frank--Wolfe iterations for temporal reconstruction
and set the ridge regularization coefficient to $\lambda=10^{-5}$
for spatial reconstruction.

\par\addvspace{1.5\baselineskip}
\noindent\textbf{Candidate-pool construction.}
We set the candidate-pool multiplier to $m=2$.
For each frame $t>1$, the minimum candidate count is $\min(N,mK_t)$,
where $N$ is the number of patch tokens in the frame and $K_t$ is its
output token budget. Spatial basis tokens count toward this minimum.
If the initial pool $\mathcal{C}_t^{(0)}$ contains fewer than
$\min(N,mK_t)$ tokens, we add unselected tokens in descending order of
temporal reconstruction error until the pool reaches the minimum.
Otherwise, we keep the initial pool unchanged. We pass the resulting
candidate pool $\mathcal{C}_t$ to the Spatial Filtering Module.

\section{Qualitative Analysis}
\label{sec:supp_qualitative}

\begin{figure*}[!t]
    \centering
    \includegraphics[width=\textwidth]{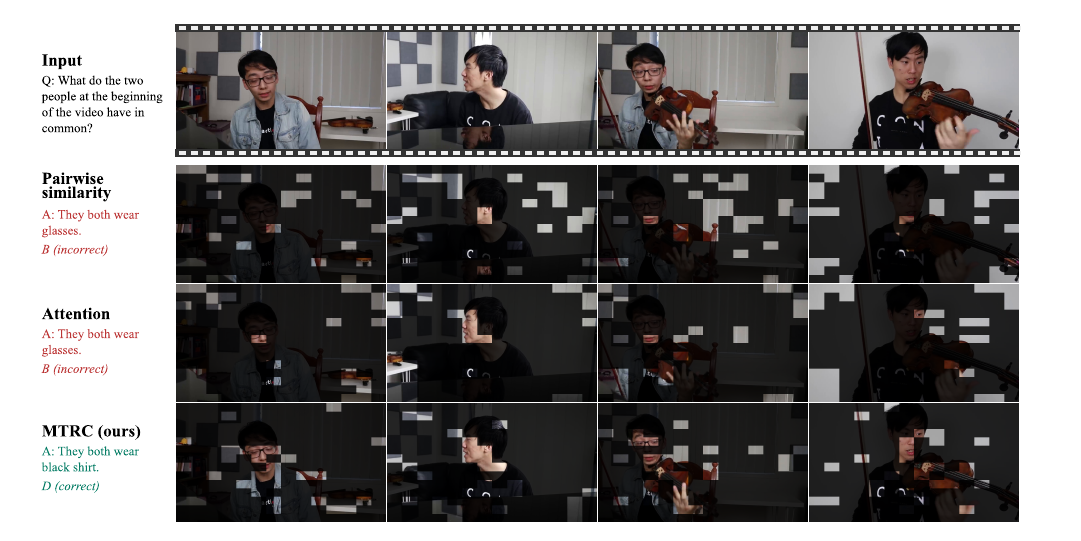}
    \caption{
    Qualitative comparison on identifying a shared attribute
    of two people in a video.
    }
    \label{fig:supp_qualitative1}
\end{figure*}

\begin{figure*}[!t]
    \centering
    \includegraphics[width=\textwidth]{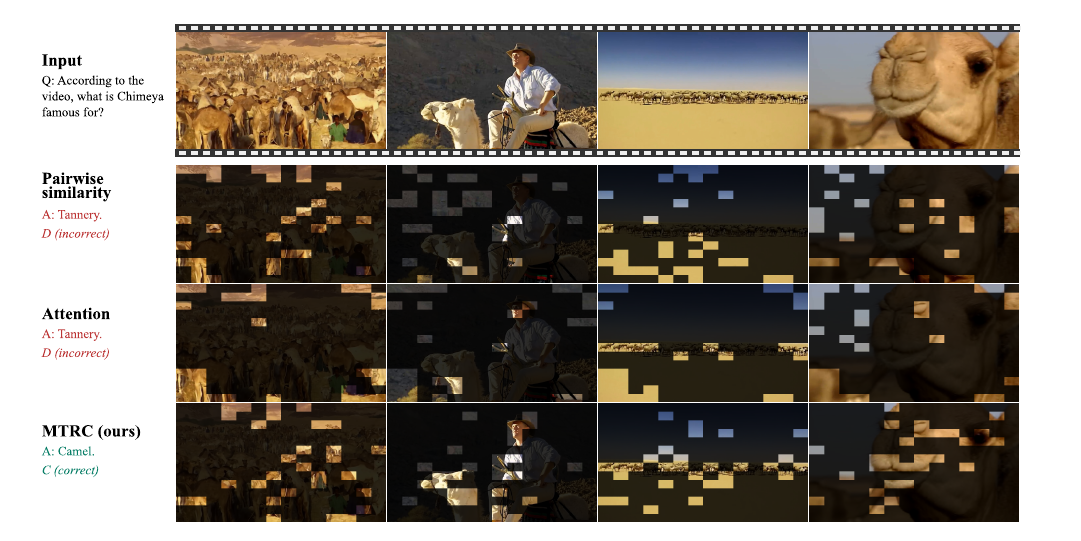}
    \caption{
    Qualitative comparison on identifying what a location
    is known for.
    }
    \label{fig:supp_qualitative2}
\end{figure*}

\begin{figure*}[!t]
    \centering
    \includegraphics[width=\textwidth]{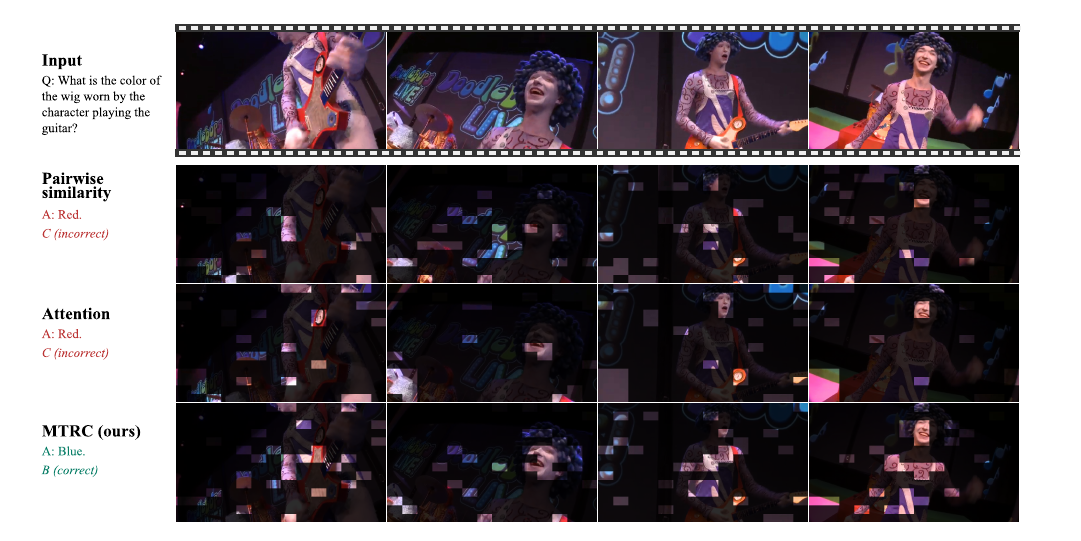}
    \caption{
    Qualitative comparison on recognizing the wig color
    of a character playing the guitar.
    }
    \label{fig:supp_qualitative3}
\end{figure*}

\begin{figure*}[!t]
    \centering
    \includegraphics[width=\textwidth]{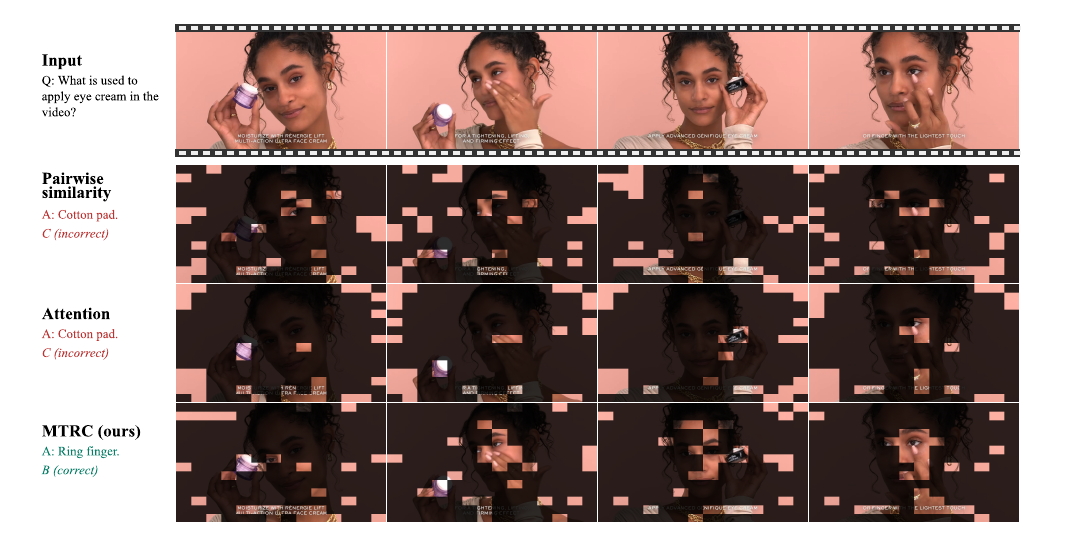}
    \caption{
    Qualitative comparison on identifying what is used
    to apply eye cream.
    }
    \label{fig:supp_qualitative4}
\end{figure*}

To further illustrate the effectiveness of MTRC, this section provides additional qualitative comparisons of token selection.
All visualization experiments are conducted on LLaVA-OneVision with a visual token retention ratio of 15\%.
As shown in Figures~\ref{fig:supp_qualitative1}--\ref{fig:supp_qualitative4}, each example presents the input frames, retained visual tokens, and the corresponding answers from the compared methods.
Bright regions indicate retained tokens, while darkened regions indicate unselected tokens.

\end{document}